\documentclass[letterpaper, 10 pt, conference]{ieeeconf}  

\IEEEoverridecommandlockouts                              

\usepackage{times}
\usepackage{color, colortbl, xcolor}

\makeatletter
\let\NAT@parse\undefined
\makeatother

\usepackage{multicol}
\usepackage{mathtools}
\usepackage[bookmarks=true]{hyperref}
\usepackage{amsfonts}
\usepackage{amsmath}
\usepackage{subcaption}
\usepackage{multicol}
\usepackage[pdftex]{graphicx}   
\usepackage[ruled,vlined,linesnumbered]{algorithm2e}
\usepackage[noend]{algpseudocode}
\usepackage{cancel}

\usepackage[textfont=md,font=footnotesize]{caption}

\let\labelindent\relax
\usepackage{enumitem}

\usepackage{multirow}

\DeclareMathOperator*{\argmin}{arg\,min}

\newcommand{\shnote}[1]%
    {\textcolor{blue}{SH: #1}}
\newcommand{\jcnote}[1]%
    {\textcolor{green}{\textbf{JC: #1}}}
\newcommand{\mgnote}[1]%
    {\textcolor{red}{\textbf{MG: #1}}}
\newcommand{\ssnote}[1]%
    {\textcolor{purple}{\textbf{SS: #1}}}

\newcommand{\dbound}{\hat{\mathcal{D}}}

\newcommand{\constraintset}{\mathcal{C}}
\newcommand{\safeset}{\mathcal{S}}

\newcommand{\tvar}{t} 
\newcommand{\tdummy}{\tau} 
\newcommand{\R}{\mathbb{R}} 
\newcommand{\ctrl}{u}
\newcommand{\ctrlsafe}{\ctrl_s^*}
\newcommand{\ctrlperf}{\ctrl_p}
\newcommand{\dstb}{d}
\newcommand{\cfunc}{u(\cdot)}
\newcommand{\dfunc}{d(\cdot)}
\newcommand{\cset}{\mathcal{U}}

\newcommand{\dset}{\mathcal{D}}

\newcommand{\state}{x}
\newcommand{\traj}{\xi} 

\newcommand{\dyn}{f} 
\newcommand{\targetfunc}{c}

\newcommand{\vfunc}{V}

\newcommand{\vconv}{\vfunc^*}
\newcommand{\warmfunc}{w}

\newcommand{\vfuncw}{\vfunc_{\warmfunc}}

\newcommand{\runningexample}[1]%
{
\textbf{Running example:}
\textit{#1}
}

\usepackage{balance}

\usepackage[%
    style=numeric-comp,
    sorting=none,
    backend=biber,
    sortcites=true,
    doi=false,
    firstinits=true,
    hyperref,
    isbn=false,
    eprint=false,
    maxcitenames=3, 
    minbibnames=3, 
    maxbibnames=4, 
    block=none]
    {biblatex}
    
    \renewbibmacro{in:}{}
    \AtEveryBibitem{%
  	\clearlist{language}%
  	\clearfield{pages}%
	}
\bibliography{references}

\title{\LARGE \bf
Scalable Learning of Safety Guarantees for Autonomous Systems using Hamilton-Jacobi Reachability}

\author{Sylvia Herbert*, Jason J. Choi*, Suvansh Sanjeev, Marsalis Gibson, Koushil Sreenath, and Claire J. Tomlin 
\thanks{*Indicates co-first authors. This research is supported by the DARPA Assured Autonomy program, the ONR BRC in Multibody systems, the SRC CONIX Center, the NSF VeHICal project and NSF grant CMMI-1931853. Contact info: \href{mailto:sherbert@ucsd.edu}{sherbert@ucsd.edu},
\href{mailto:jason.choi@berkeley.edu}{jason.choi@berkeley.edu},
\href{mailto:sqs@cs.cmu.edu}{sqs@cs.cmu.edu},
\href{mailto:mtgibson@berkeley.edu}{mtgibson@berkeley.edu},
\href{mailto:koushils@berkeley.edu}{koushils@berkeley.edu},
\href{mailto:tomlin@berkeley.edu}{tomlin@berkeley.edu}}
}

\begin{document}

\maketitle
\thispagestyle{empty}
\pagestyle{empty}

\begin{abstract}
Autonomous systems like aircraft and assistive robots often operate in scenarios where guaranteeing safety is critical. Methods like Hamilton-Jacobi reachability can provide guaranteed safe sets and controllers for such systems. However, often these same scenarios have unknown or uncertain environments, system dynamics, or predictions of other agents. As the system is operating, it may learn new knowledge about these uncertainties and should therefore update its safety analysis accordingly. However, work to learn and update safety analysis is limited to small systems of about two dimensions due to the computational complexity of the analysis. In this paper we synthesize several techniques to speed up computation: decomposition, warm-starting, and adaptive grids. Using this new framework we can update safe sets by one or more orders of magnitude faster than prior work, making this technique practical for many realistic systems. We demonstrate our results on simulated 2D and 10D near-hover quadcopters operating in a windy environment.

\end{abstract}

\section{Introduction}
\label{sec:introduction}

Safety-critical scenarios are situations in which autonomous systems must be able to ensure safety during operation. Many techniques have been studied to produce safe controllers that will keep a particular system within a guaranteed safe set. However, in the real world, there will inevitably be unexpected changes in the system or environment that may violate initial assumptions, invalidating safety guarantees. It is therefore crucial that the system is able to react to changing knowledge and to update its safety controllers and guarantees accordingly.
 
Safe learning for dynamical systems is an increasingly active research area.
A few researchers take the approach of \textit{safe (machine) learning}, where learning algorithms are updated or guided to provide safer resulting controllers during training. This can be done by, for example, projecting an algorithm's update of a policy to a valid constraint set \cite{achiam2017constrained}, moving slowly in uncertain areas \cite{kahn2017uncertainty}, or using Lyapunov functions to drive the learning of a safe policy \cite{ chow2018lyapunov}. 

One way to enable any reinforcement algorithm to maintain safety is to precompute a fixed guaranteed safe set and safety override controller to keep the system within that set. This can be done using techniques like control barrier functions \cite{cbfsurvey} or Hamilton-Jacobi (HJ) reachability \cite{bansal2017hamilton}. 

Another line of work takes the approach of \textit{learning for safety.} The premise of this work is that a safe policy or safe set will not be valid if the assumptions about the system model or constraints no longer hold. Therefore, the system must update its assumptions and corresponding safe methods as the system learns more about the environment \cite{richards2018lyapunov, akametalu2015temporal,  marco2020excursion, bajcsy2019efficient, kolter2019learning, beckers2020safe, choi2020reinforcement}.

Finally, \textit{learning for safety }and \textit{safe (machine) learning} can be combined by jointly producing a guaranteed safe set and corresponding safe controller based on information gathered online, while simultaneously learning a performance controller \cite{akametalu2014safelearning, fisac2019safelearning,berkenkamp2017safe,cheng2019end, rosolia2017learning,dean2019safely,aswani2013provably, huh2020safe}. These approaches produce strong results for small (2-3D) systems, but struggle to extend to higher-dimensional systems due to the computational complexity of updating the safe set and safety controller.  For example, the safe-learning framework in  \cite{akametalu2014safelearning,fisac2019safelearning} uses HJ reachability wherein the state space is discretized and scales as $O(N^D)$, where $N$ is the number of grid points in each dimension and $D$ is the number of dimensions.

 In this work we seek to efficiently compute safe controllers and sets \textit{online} for realistic systems without compromising on strong theoretical guarantees. To accomplish this, we build upon \cite{akametalu2014safelearning,fisac2019safelearning} by incorporating three methods to improve computation: 
 \begin{enumerate}[leftmargin=2em]
     \item Decomposing dynamical systems with ``self-contained subsystems'' to improve computation by orders of magnitude while maintaining exact results \cite{Chen2016DecouplingJournal}.
     \item When new information is learned about the system or environments, updating the safe set directly using warm-starting rather than completely recomputing the safe set. Recent work \cite{herbert2019reachability,bajcsy2019efficient} proved that this will converge to exact or conservative results while reducing iterations to convergence of the computation.
     \item Initializing the safe set computation with a coarse grid that is refined over time, while maintaining exact or conservative guarantees.
 \end{enumerate}

 We demonstrate the new learning for safety framework on a 2D quadcopter model and a 10D near-hover quadcopter model experiencing unknown wind disturbances. Due to the exponential computational scaling, computing safe sets for the 10D model using HJ Reachability directly would be intractable. Decomposing the system into one 2D and two 4D subsystems using \cite{Chen2016DecouplingJournal} makes the computation tractable at 2 to 3 hours. Further incorporating the warm-starting and adaptive grid reduces the time further to an average of 3.3 minutes. Simulation demonstrations in the Robot Operating System (ROS) \cite{quigley2009ros} environment show the quadcopters maintaining and updating safety analyses online, with an average safety update time of less than one second for the 2D system and 206.6 seconds for the 10D system.

\section{Background: Safe Learning}
\label{sec:original}

The purpose of the prior framework in \cite{akametalu2014safelearning,fisac2019safelearning} is to (a) provide a safe set within which the system can freely learn a performance controller, and (b) update that safe set based on learned information about the system or environment.  Note that there are two forms of learning happening: \textit{safe learning} of the performance controller, and \textit{learning for safety} by updating the safe set based on learned parameters.  This section provides background on the existing safe learning framework that our work builds on. 

\subsection{Computing Safe Sets \& Controllers using HJ Reachability \label{subsec:initial_safety_computation}}
\vspace{-1em}
Consider a general system $\dot{\state} = \dyn(\state, \ctrl, \dstb)$ describing the evolution over time of the state $\state\in\mathbb{R}^n$ under system control inputs $\ctrl\in\cset$ in the presence of environmental disturbances $\dstb\in\dset(\state)$, which may vary with the state. Let there be a state constraint set $\constraintset\subset\R^n$ in the environment (representing, for example, obstacle boundaries or bounds on velocity of the system).
The maximal safe control-invariant set is denoted as $\safeset(\constraintset,\dset)$, with corresponding optimal safety controller, $\ctrl^*(\state)$, using which the system is guaranteed to remain in the constraint set $\constraintset$ even when experiencing worst-case disturbances. One way to compute this set is through HJ reachability analysis. Solving for both $\safeset(\constraintset,\dset)$ and $\ctrl^*(\state)$ can be posed as an optimal control problem whose value is \vspace{-0.5em}
\begin{equation} \label{eq:value_function}
\vfunc(\state) = \sup_{\dfunc} \inf_{\cfunc} \max_{\tdummy \in [\tvar,0]} \targetfunc\Big(\traj(\tdummy; \state, \tvar, \cfunc, \dfunc)\Big). \vspace{-0.5em}
\end{equation}
Here, $\traj(\tdummy; \state, \tvar, \cfunc, \dfunc)$ denotes the state reached by the system $f$ at timestep $\tdummy$ when starting at state $\state$ and time $\tvar$. 
The cost, $\targetfunc(\state)$, is the signed distance function for the constraint set $\constraintset$, such that $\constraintset=\{\state:\targetfunc(\state) \leq 0\}$. 

By formulating the value function with a maximum over time, the analysis captures whether a trajectory ever violates the state constraints. The infimum over control ensures that the control input will act optimally to keep the system within the constraint set $\constraintset$, and the supremum over disturbance assumes that the disturbances will be acting in an optimally adversarial manner.

The value function optimization \eqref{eq:value_function} is in general non-convex and challenging to compute. One method is to use dynamic programming: the value of the function at the final time is set as equal to the cost $\vfunc(\state,0) = \targetfunc(\state)$, and then iterated backwards in time using the Hamilton-Jacobi-Isaacs Variational Inequality (HJI VI) \cite{fisac2015reach} until convergence:\vspace{-.5em}
\begin{equation}
\begin{aligned}
    \label{eq:HJIVI}
    &0 =\max\Big\{
    \targetfunc(\state)-\vfunc(\state,\tvar), \\
    &D_\tvar \vfunc(\state,\tvar)+
    \min_\ctrl \max_\dstb  \langle \nabla \vfunc(\state,\tvar), \dyn(\state,\ctrl,\dstb) \rangle \Big\}. \vspace{-0.5em}
    \end{aligned}
\end{equation}

In the infinite horizon scenario, we drop the dependence on time and denote the optimal converged value function as $\vconv(\state)$ with corresponding safe set $\safeset(\constraintset,\dset) = \{ \state : \vconv(\state)\leq 0\}$. This set captures states from which optimal trajectories of the system maintain non-positive cost over an infinite time horizon, and therefore never violate the constraint set despite worst-case disturbances.  The gradients of this function can be used to compute the optimal safety control, 
$    \ctrlsafe(\state) = \argmin_\ctrl \max_\dstb  \langle \nabla \vconv(\state), \dyn(\state,\ctrl,\dstb) \rangle.$

Every non-positive level set of the value function provides a safe invariant set. We therefore set the default safe set as the subzero level set of the converged value function.

\subsection{Safe Learning: Learning the Performance Controller within the Safe Set\label{subsec:safe_learning}}
If the system is within the safe set, it is free to explore and learn a desired policy or performance controller, denoted $\ctrlperf(\state)$. As the system approaches the boundary of the safe set, the optimal safety controller $\ctrlsafe(\state)$ overrides the learned policy to keep the system within the safe set. By providing a safe set and safety controller, the learning process can happen confidently without concern of safety violations.
Due to the switching between the performance and safety controllers, this structure is more amenable to \textit{off-policy} learning methods that do not rely on the assumption that the data they are trained on is collected from the method's controller itself. 

\subsection{Learning for Safety: Updating the Safe Set and Controller\label{subsec:learning_safety}}
The initial safe set from Sec.~\ref{subsec:initial_safety_computation} was computed based on certain assumptions about the system and the environment. These assumptions may change in light of data the system collects online.
If this happens, the safety guarantees established under the initial assumptions will no longer hold, which will also make the optimal safety controller invalid. Therefore, the safe-learning framework must include data-driven methods to learn about the environment in real time and update the corresponding safety guarantees.

We assume that uncertainties in the dynamics can be described as disturbances to the system (e.g. wind, model-plant mismatch). The system measures these state($x_j$)-dependent disturbances($d_j$) while exploring the state space. A collection of these measurements $\{(x_j, \hat{d_j})\}_{j=1}^{N}$are then used to estimate the bounds of disturbance across the state space using Gaussian Process (GP) regression \cite{rasmussen2003gaussian}.
GP regression is suitable for our problem since it can capture both epistemic uncertainty (due to limited data) and the system's inherent stochasticity (such as wind effect). Moreover, we can include measurement noise in the formulation, which allow us to account for estimation errors in the disturbance measurement.

Based on the collected data, GP regression is performed by optimizing the kernel parameters and taking the posterior distribution of GP conditioned on the data. We refer to \cite{rasmussen2003gaussian} for more details. The output is a $\delta$-confidence interval of the gaussian distribution from the GP model which approximates the updated disturbance bound $\dset(\state)$ across the state space. 
The safe set computed based on this disturbance bound will provide a high-probability safety guarantee under physical disturbance, model-plant mismatch, and disturbance estimation errors.

Once the new disturbance bounds are learned, the safe set must be updated to reflect these changed assumptions.  In the prior safe-learning framework \cite{akametalu2014safelearning,fisac2019safelearning}, the update occurs simply by recomputing the entire safe set, as in Sec.~\ref{subsec:initial_safety_computation}.  While recomputation occurs, the system finds and stays within a negative sublevel set of the previous safe value function where the disturbance assumptions still hold.  This contraction tends to be overly conservative, but provides a temporary safe region to stay within while the new safe set is computed.
Unfortunately, recomputing the safe set online using the prior framework is infeasible for most realistic systems due to the computational complexity of HJ reachability.

\section{Accelerated Learning for Safety}
\label{sec:new_framework}

The prior framework \cite{akametalu2014safelearning,fisac2019safelearning} was demonstrated on a 2D system due to its issues with computational scalability. We propose modifications to the safe set computation in order to handle higher-dimensional systems.

\subsection{Incorporating Decomposition}
Some higher-dimensional systems can be decomposed into smaller subsystems that can be analyzed independently \cite{Chen2016DecouplingJournal,Chen2016DecouplingExact}. When possible, this reduces the computation time by potentially orders of magnitude. This is simply because splitting the analysis into multiple computations with lower dimensions changes the magnitude of the exponential scaling. For example, a 10D quadcopter model that will be used in Sec.~\ref{sec:computation_results} can be decomposed into two 4D and one 2D system, changing the computational complexity from $O(N^{10})$ to $O(N^4+N^4+N^2)$, where $N$ is the number of grid points in each dimension.

In order to provide \textit{exact guarantees} on the safe set and controller computation while using decomposition, the dynamic system must be either completely decoupled or coupled via \textit{self-contained subsystems} (details in \cite{Chen2016DecouplingJournal,Chen2016DecouplingExact}).  For systems that do not have these properties, the remaining two components of the updated safe learning framework in Sec.~\ref{subsec:warm-start} and Sec.~\ref{subsec:coarse} still apply.

\subsection{Incorporating Warm-Starting \label{subsec:warm-start}}

Standard HJ reachability analysis requires that any computation must be initialized at the terminal cost function, i.e. $\vfunc(\state,0)=\targetfunc(\state)$, in order to provide guarantees.  In contrast, the reinforcement learning community often employs the technique of warm-starting, wherein a ``best guess'' initialization is used, and therefore the computation may converge in fewer iterations (if convergence can be achieved).  Recent research was able to prove \textit{exact or convervative convergence} of warm-started HJ reachability analyses \cite{akametalu2018minimum,herbert2019reachability}.

We can employ this proven technique to the learning for safety framework. Upon changes in information or assumptions (e.g. change in disturbances, control authority, obstacles, model parameters), one can initialize a new computation using the previously computed safe set, rather than reinitializing from the constraint set.  By warm-starting from a previous solution that was based on slightly different assumptions, \cite{herbert2019reachability} shows empirically that the computation will generally converge in fewer iterations.

We initialize with the previously computed value function, which we will define as $\warmfunc(\state)$. 
The new value function is computed using the standard HJI VI \eqref{eq:HJIVI}, with initialization $\vfuncw(\state,0) = \warmfunc(\state)$, constraint set $\constraintset$, and updated disturbance bounds $\dset(\state)$.  The computation is run until convergence, outputting the converged value function $\vconv_\warmfunc(\state)$ and safe set $\safeset(\constraintset,\dset)=\{\state:\vconv_\warmfunc(\state)\leq 0\}$.

\subsection{Incorporating Coarse Approximations\label{subsec:coarse}}

The last component of the new safe-learning framework extends on the work in \cite{herbert2019reachability} by applying the warm-starting technique more generally.  In addition to warm-starting from previous solutions, one could warm-start from coarse approximations to the safe set. Through an initial computation using a coarse grid and cheap gradient approximations, a very quick rough approximation of the true safe set can be computed.  This approximation can then be refined by using it as an initialization to a fine, higher-accuracy computation. Adaptive grids are used in the fluid mechanics and reinforcement learning communities \cite{moore1995parti,macneice2000paramesh,berger1989local,du2020multi}, and with the warm-starting convergence proofs these techniques can now be applied to reachability analysis.

We initialize again with the previously computed value function, this time over a coarse grid $\warmfunc^{\text{coarse}}(\state)$. 
The new value function is computed as in Sec.~\ref{subsec:warm-start}, outputting the converged value function $\vfuncw^{*,\text{coarse}}(\state)$. This is then used to initialize a second computation over a coarse grid, $\vfuncw^{\text{fine}}(\state,0) = \vfuncw^{*,\text{coarse}}(\state)$. The final converged value function is denoted as $\vconv_\warmfunc(\state) = \vfuncw^{*,\text{fine}}(\state)$.

\section{Computation Comparison}
\label{sec:computation_results}

We first isolate the safe set computation component of the framework and perform a computational comparison to the prior work in \cite{akametalu2014safelearning,fisac2019safelearning}. The computation comparison will be for a hypothetical experiment in which an initial safe set is computed for a 10D near-hover quadcopter \cite{Bouffard12}, and then must be updated when the GP produces a new disturbance bound estimate. The 10D model has states $( p_x,  p_y,  p_z)$ denoting the position, $( v_x,  v_y,  v_z)$ for velocity, $(\theta_x, \theta_y)$ for pitch and roll, and $(\omega_x, \omega_y)$ for pitch and roll rates. Its controls are the desired pitch and roll angle $(S_x, S_y)$, and vertical thrust $T_z$. The disturbances $(\dstb_x, \dstb_y, \dstb_z)$ represent wind, and $g$ is gravity. Its model is:\vspace{-.5em}

\begin{equation}
\label{eq:Quad10D_dyn}
\begin{aligned}
\footnotesize
\begin{array}{c}
	\left[
	\begin{array}{c}
	\dot p_x\\
	\dot v_x\\
	\dot\theta_x\\
	\dot\omega_x\\
	\dot p_y\\
	\dot v_y\\
	\dot\theta_y\\
	\dot\omega_y\\
	\dot p_z\\
	\dot v_z
	\end{array}
	\right]
	=
	\left[
	\begin{array}{c}
	 v_x\\
	g \tan \theta_x  + \dstb_x\\
	-d_1 \theta_x + \omega_x\\
	-d_0 \theta_x + n_0 S_x\\
	 v_y\\
	g \tan \theta_y  + \dstb_y\\
	-d_1 \theta_y + \omega_y\\
	-d_0 \theta_y + n_0 S_y\\
	 v_z\\
	(k_T/m) T_z - g  + \dstb_z 
	\end{array}
	\right].
\end{array}\\
\end{aligned}
\end{equation}

By limiting the quadcopter to near-hover (small pitch and roll) conditions, there is assumed to be no coupling through yaw. The parameters $d_0, d_1, n_0, k_T$, and the control bounds $\cset$ that we used were $d_0 = 10, d_1 = 8, n_0 = 10, k_T = 4.55, |S_x|, |S_y| \le 14^{\circ}, 0.6mg \le T_z \le 1.4mg$. 
Extending the framework to use this model is not an easy task: updating the safe set online for a 10D system using the original safe-learning framework would take $O(N^{10})$, where N is the number of grid points. With $50$ grid points in each dimension and an estimated time of $0.001$s for each grid point to reach convergence, this computation would be intractable.

We assume an initial disturbance bound with mean $\mu=0~\text{m/s}^2$ and variance $\sigma^2 = 0.01~\text{m/s}^2$, bounded at $99.7\% (3\sigma)$ confidence. The disturbance bound is then updated by GP regression from disturbance data collected in simulation (Sec. \ref{subsec:10DDemo}) under wind effect, and an updated computation must occur to incorporate this new information into the safe set.  All computation in this section is done using a 2015 MacBook Pro with a 2.8 GHz Quad-Core Intel processor and 16GB of memory. All computation results can be seen in Table~\ref{tab:computation_comparison}.

\subsection{Incorporating Decomposition}
We first use the decomposition technique from \cite{Chen2016DecouplingJournal} to decompose the 10D dynamical system into three subsystems in the $x$ (4D), $y$ (4D), and $z$ (2D) dimensions.  This reduces the number of grid points to iterate over from $9.76$E$16$ for the 10D system to $5.76$E$6$ for the 4D systems and $1$E$4$ for the 2D system. The safe set can be computed exactly in each subsystem with the corresponding safety controllers for the $x, y,$ and $z$ subsystems separately.

\begin{figure}
\centering
\includegraphics[width=.8\columnwidth]{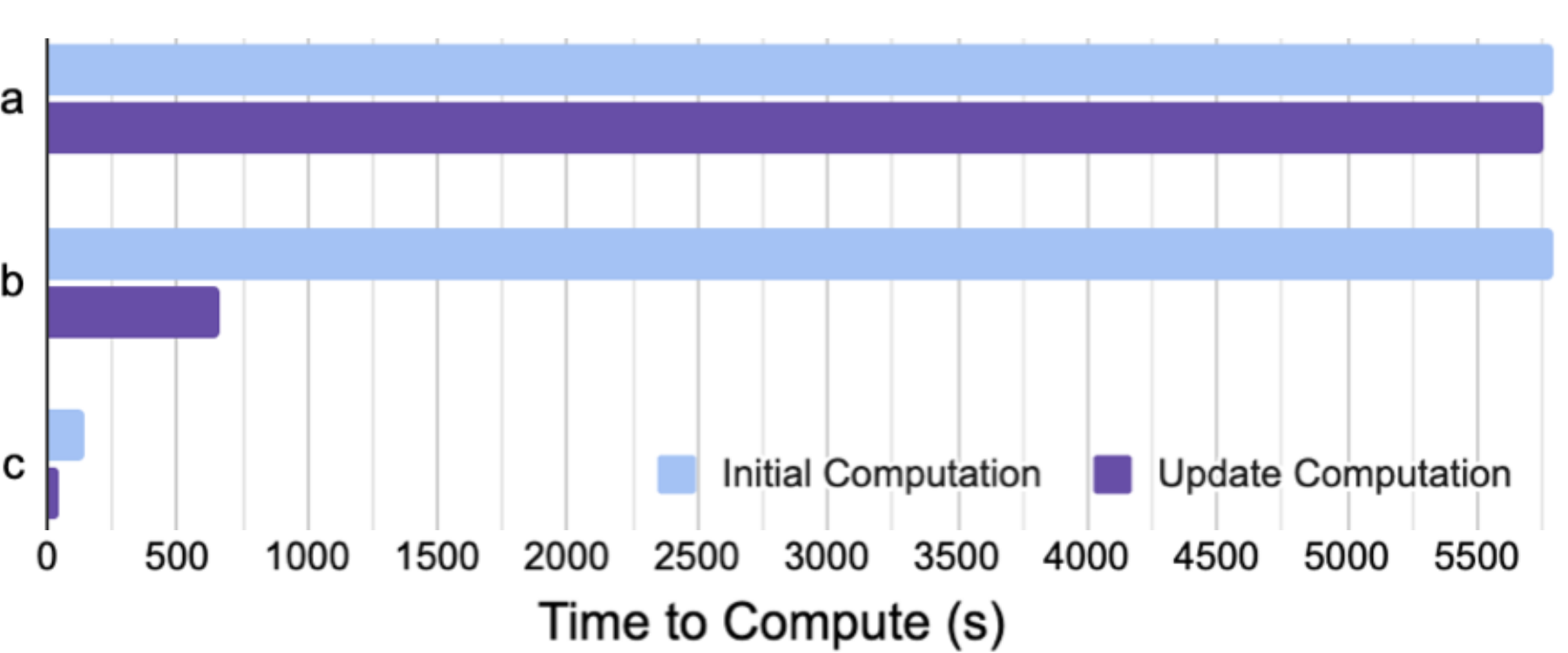}
\caption{Computation time comparison for 4D x-subsystem of 10D quadcopter model. Blue columns show the compute time for the initial computation, and purple columns show the compute time for the updated computation with new disturbance bound from data. The sets of bars show the computation when incorporating (a) decomposition, (b) decomposition and warm-starting, (c) decomposition and warm-starting and coarse initializations. Using the new framework (c), the computation of the 4D subsystem takes minutes instead of hours, making updating a 10D safe set online tractable for the first time.}
\label{fig:computation_comparison}
\vspace{-.5em}
\end{figure}

Using the decomposed system reduces the computation time from being intractable to the order of hours. Fig.~\ref{fig:computation_comparison}a shows the computation time for the 4D x-subsystem for the initial computation (blue, 5797s or 1.6 hours) and updated computation (purple, 5752s or 1.6 hours). Despite these improvements, this time frame is still not sufficiently efficient for the online computation for safe sets for real experiments.

\begin{table}\scriptsize
\caption {Time Comparison of Computation Methods for 4D Subsystem of 10D Quadcopter Model} 
\begin{tabular}{cc|c|c|c|}
\cline{3-5}
                                                                                              &                & \begin{tabular}[c]{@{}c@{}}Prior \\ Framework \\ (with   decomp.)\end{tabular} & \begin{tabular}[c]{@{}c@{}}Decomp. +\\ Warm-Start\end{tabular}          & \begin{tabular}[c]{@{}c@{}}New Framework\\ (Decomp. + Warm-\\Start + Coarse Init.)\end{tabular} \\ \hline
\multicolumn{1}{|c|}{\multirow{4}{*}{\begin{tabular}[c]{@{}c@{}}Initial\\ Comp\end{tabular}}} & x (4D)         & 5797 s                                                                        & 5797 s                                                                & 143.6 s                                                                                           \\ \cline{2-5} 
\multicolumn{1}{|c|}{}                                                                        & y (4D)         & 5645 s                                                                        & 5645 s                                                                & 145.6 s                                                                                            \\ \cline{2-5} 
\multicolumn{1}{|c|}{}                                                                        & z (2D)         & 1.7 s                                                                           & 1.7 s                                                                   & 1.4 s                                                                                             \\ \cline{2-5} 
\multicolumn{1}{|c|}{}                                                                        & \textbf{total} & \textbf{\begin{tabular}[c]{@{}c@{}}9656s (2.7hr)\end{tabular}}         & \textbf{\begin{tabular}[c]{@{}c@{}}9656s (2.7hr)\end{tabular}} & \textbf{\begin{tabular}[c]{@{}c@{}}290.6s (4.8min)\end{tabular}}                          \\ \hline
\multicolumn{1}{|c|}{\multirow{4}{*}{\begin{tabular}[c]{@{}c@{}}Update\\ Comp\end{tabular}}}  & x (4D)         & 5752 s                                                                             & 659.4 s                                                                     & 42.3 s                                                                                               \\ \cline{2-5} 
\multicolumn{1}{|c|}{}                                                                        & y (4D)         & 5535 s                                                                             & 667.3 s                                                                     & 80.3 s                                                                                               \\ \cline{2-5} 
\multicolumn{1}{|c|}{}                                                                        & z (2D)         & 1.1 s                                                                             & 0.4 s                                                                     & 0.3 s                                                                                               \\ \cline{2-5} 
\multicolumn{1}{|c|}{}                                                                        & \textbf{total} & \textbf{\begin{tabular}[c]{@{}c@{}}11288s (3.1hr)\end{tabular}}                  & \textbf{\begin{tabular}[c]{@{}c@{}}1327s (22min)\end{tabular}}          & \textbf{\begin{tabular}[c]{@{}c@{}}123s (2min)\end{tabular}}                                    \\ \hline 
\end{tabular}\vspace{-2.5em}
\label{tab:computation_comparison}
\end{table}

\subsection{Incorporating Warm-Starting}
When the disturbance bounds change, we can warm-start the update from the previously computed safe set. Computation results for decomposition with warm-starting are shown in Fig.~\ref{fig:computation_comparison}b.  This method does not impact the time to compute the initial safe set shown in blue at 5797s (1.6 hours), but does impact the time to \textit{update} the safe set shown in purple at 659.4s (11 minutes).

\subsection{Incorporating Coarse Approximations}
By initializing computations with a coarse grid that is run to convergence, broad changes in the safe set can be computed quickly. The resulting value function can then be migrated to a finer grid to refine the function more accurately until convergence.  Fig.~\ref{fig:computation_comparison}c illustrates the computational results for our new framework that synthesizes this coarse approximation along with decomposition and warm-starting. The initial computation of the 4D x-subsystem using the new framework takes 143.6s, or 2.4min (blue). The updated computation takes 42.3s (purple).
Using all three new components reduces computation by an order of magnitude compared to the decomposed system, and by several orders of magnitude compared to the full 10D system! 

\subsection{The New Safe-Learning Framework}
This updated \textit{learning for safety} framework reduces the computation of the 10D set from an intractable length of time to 4.8 minutes when computed serially, or 2.4 minutes in parallel.  When updating from a previous solution, the 10D computation is 2 minutes in series and 1.3 minutes in parallel. Note that these computations may take moderately longer when the computer is also learning the performance controller and running a simulator.
\begin{figure}
\centering
\includegraphics[width=.45\columnwidth]{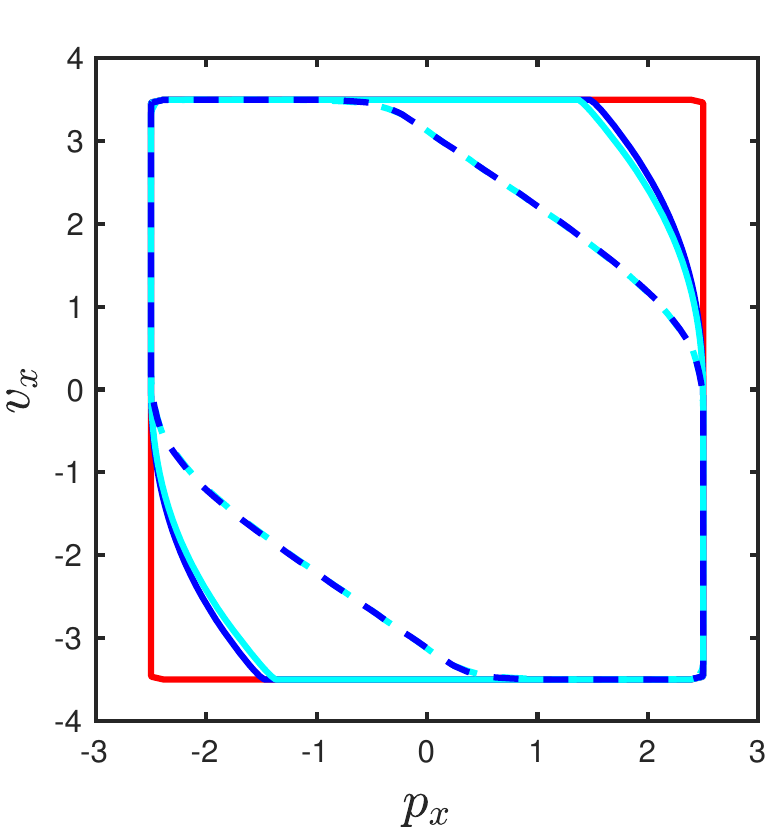}
\caption{Solid red line: constraint set boundary of 4D x-subsystem of 10D quadcopter model projected to position and velocity dimensions. Solid lines: initial safe set $\safeset(\constraintset,\dset)$ boundary computed using prior framework with decomposition (dark blue, 5797 seconds) and the new framework (cyan, 143.6 seconds). Dashed lines: updated safe set with new disturbances computed using prior framework with decomposition (dark blue, 5752 seconds) and the new framework (42.3 seconds).
}
\label{fig:warmstart_comparison_4D}
\vspace{-2em}
\end{figure}

\begin{figure*}
\centering
\includegraphics[width=2\columnwidth]{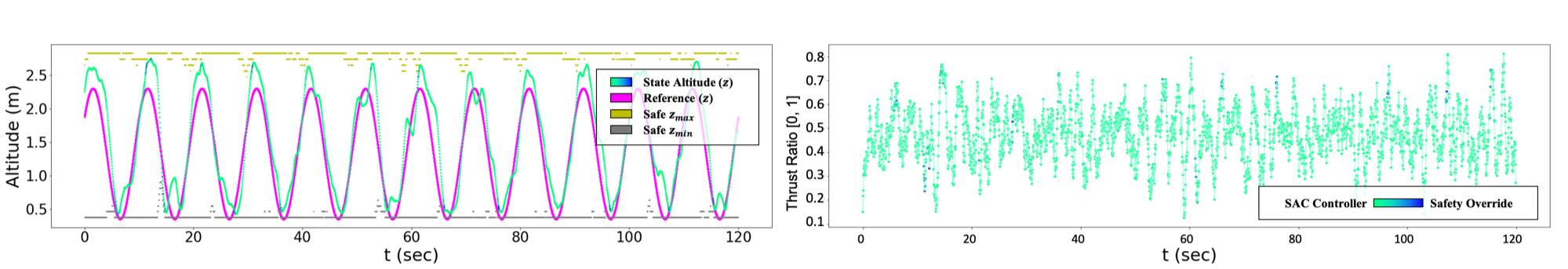}
\caption{Close-up of the learning process of the 2D quadcopter over a short time period. \textbf{Left: }Altitude of quadcopter (green), reference trajectory (magenta) and $z$-boundaries of the safe set (gray and yellow) are plotted over a snapshot of the training process. The safe set boundaries are computed as a function of the drone's velocity, explaining the narrower safe regions during periods of faster motion. \textbf{Right: }The safety controller must override the system (blue regions of green trace) whenever the quadcopter reaches the boundaries of the safe set as it learns to track the reference trajectory.}
\label{fig:vf_train_track}
\vspace{-1em}
\end{figure*}

\begin{figure*}
\centering
\includegraphics[width=1.6\columnwidth]{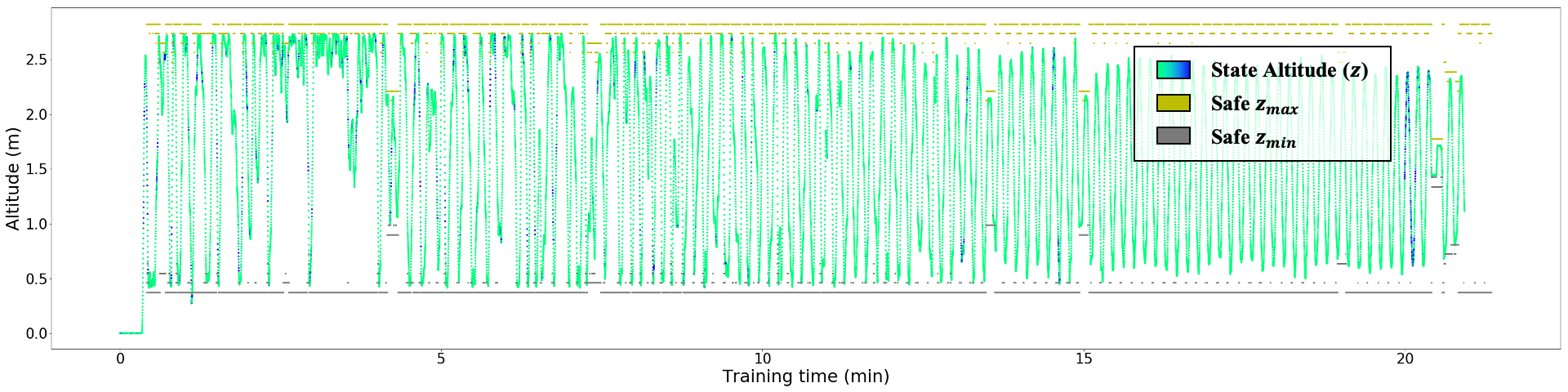}
\caption{Full view of the training process over time. Altitude of quadcopter (green) and $z$-boundaries of the safe set (gray and yellow) are plotted over the course of training. At the beginning, the quadcopter does not track the trajectory well and relies on the safety override (blue regions of green trace) and as learning progresses, the drone improves its tracking skill. Note that the safe set occasionally contracts due to noise in the disturbance estimation.}
\label{fig:vf_train_alt}
\vspace{-1.5em}
\end{figure*}

\begin{figure}[!htb]
\begin{subfigure}{\columnwidth}
  \centering
  \includegraphics[width=\linewidth]{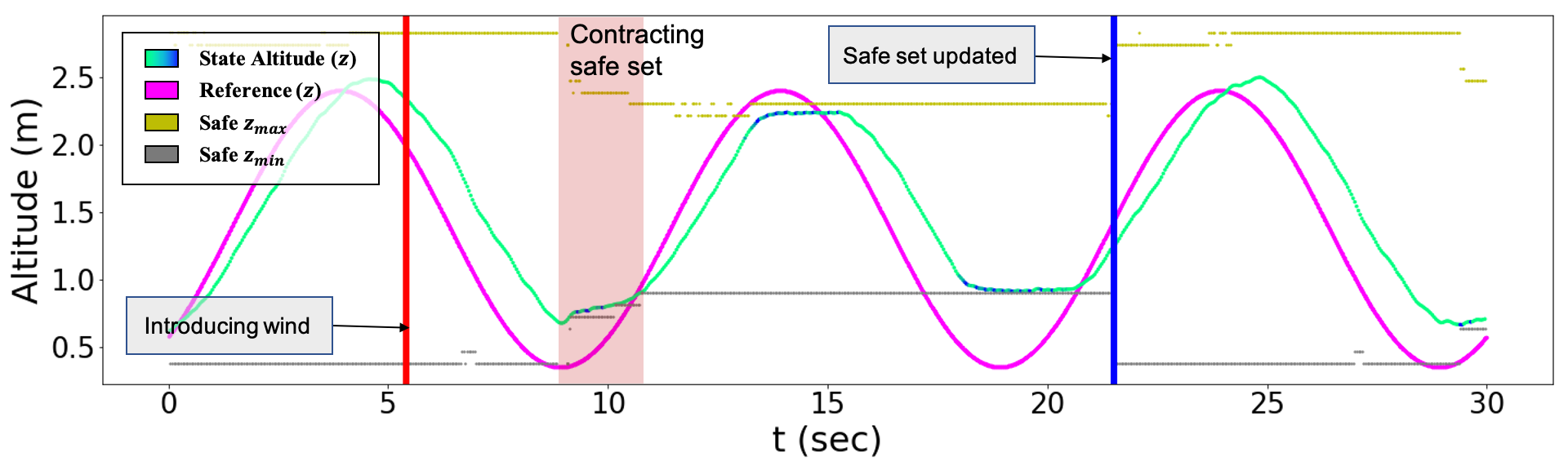}
\end{subfigure}
\begin{subfigure}{\columnwidth}
  \centering
  \includegraphics[width=\linewidth]{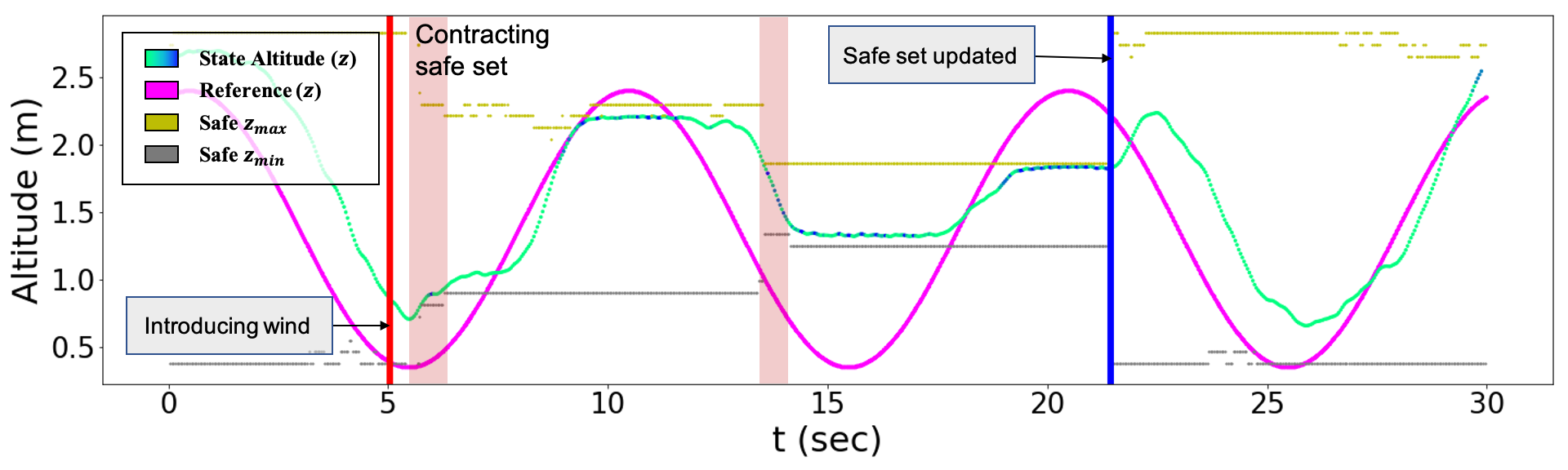}  
\end{subfigure}
\caption{Altitude of quadcopter (green), reference trajectory (magenta), $z$-boundaries of the safe set (gray and yellow). Once an external wind disturbance is introduced (red vertical line), measured disturbance is outside the confidence bound of the GP model, resulting in a contraction of the safe set (red shaded regions), allowing the quadcopter to remain safe in a conservative region while it updates the GP to reflect its belief in the disturbance function, and computes a corresponding safe set (blue vertical line). This computation is done with (top) and without (bottom) warm-starting with the old reachable set. In both cases, after the safe set is updated, the drone is allowed to expand its safe set again. \vspace{-2em}}
\label{fig:verticalflight}
\end{figure}

Figure \ref{fig:warmstart_comparison_4D} shows the updated safe set of the 4D x-subsystem projected onto the position and velocity dimensions.  The constraint (red) has bounds on position ($p_x \in [-2.5, 2.5]$m) and velocity ($v_x \in [-3.5, 3.5]$m/s).  Because computing the safe set is intractable for the 10D system, we instead show as ground truth the boundary of the safe set of the decomposed 4D x-subsystem computed with the original framework (solid blue line). This computation took 92.3 minutes. The dashed cyan line shows the boundary of the safe set updated using the new framework, which took 1.2 minutes.


\section{Simulation Demonstration}\vspace{-.1em}
\label{sec:examples}

We evaluate our framework in a Crazyflie 2.0 \cite{giernacki2017crazyflie} simulation environment \cite{dfk2017crazyflie} built on ROS \cite{quigley2009ros}. To make the simulation as realistic as possible, we build every task of the framework, which consist of state estimation, collecting disturbance data, training GP model, updating safe set, learning-based controller, and safety verification, run in parallel. The only difference from the real experiment setup is that we assume that we receive accurate positions of the drone from the Motion Capture system.

\subsection{Comparison to Original 2D Quadcopter}

The prior framework \cite{akametalu2014safelearning,fisac2019safelearning} focused on ensuring the safety of a 2D quadcopter that is learning to follow a simple sinusoidal reference trajectory.  For proper comparison we will recreate this experiment. The quadcopter is represented by a 2D affine model:\vspace{-0.5em}
\begin{equation}
    \dot{\state}_1 = \state_2, \quad
    \dot{\state}_2 = k_T \ctrl + g + k_0 + \dstb(\state) \vspace{-0.5em}
\end{equation}
\noindent where $\state_1$ is the height, $\state_2$ is vertical velocity, and $\ctrl \in [0,1]$ is the normalized motor thrust command.  The parameters $k_T$ and $k_0$ are specific to the quadcopter, and the gravity is $g = -9.8~\text{m/s}^2$. The disturbance $\dstb(\state)$ represents unmodeled forces in the system that affect the acceleration of the system (e.g. external wind). In the simulation environment, the wind blowing from a fan on the ground is emulated as a vertical acceleration applied to the drone sampled from a gaussian distribution, whose mean and variance is a function of the altitude. Their values are maximal near the ground and diminish to zero when the altitude gets higher.

The task of the quadcopter is to follow a vertical reference trajectory without getting too close to the ground or ceiling, thus the state constraints are set as $\constraintset = \{\state : 0.35~\text{m} \leq \state_1 \leq 2.8~\text{m}\}$.  The vertical reference trajectory cycles between an altitude of $0.35~$m and $2.3~$m.

The performance controller is adapted from part of a public code repository \cite{VitchyrRLkit} implementing the Soft-Actor Critic (SAC) algorithm \cite{Haarnoja18}, an off-policy reinforcement learning method that jointly learns a Q-value function over state-control pairs and a maximum-entropy policy that maximizes the Q-value function. The reward function used is based on a squared error between the reference and true states, and on the difference in the heading of the trajectories. The controller and Q-value functions are represented by three-layer neural networks with 32 units per hidden layer.

The state-dependent bound $\dbound(\state)$ that the disturbance $\dstb(\state)$ may lie within is initialized to be 
$-0.3~\text{m/s}^2\leq \dstb \leq 0.3\text{m/s}^2$.
This bound is then updated by learning the Gaussian process model and setting $\dbound(\state)$ to the marginal $99.7\% (\pm3\sigma)$ confidence interval at each value of $\state$.

 In the prior work, the iterative safety re-computation  \cite{akametalu2014safelearning,fisac2019safelearning} was only used when the quadrotor was exposed to minimal disturbances.  When adding a significant disturbance (in the form of a wind from the fan), the quadcopter did not recompute the safe set, instead using a reactive contraction method to a smaller subset of the original safe set, providing a fast but overly conservative safety bound.  Here we extend their experiment to recomputing the safe set in the face of new disturbances, and will compare this to the new framework that \textit{updates} rather than recomputes the safe set.

First, we execute the SAC controller and train its policy. Before it is trained, the SAC controller would show aggressive behavior and therefore, without the safe learning framework, it is impossible to train it safely online. Figure~\ref{fig:vf_train_track}, \ref{fig:vf_train_alt} show the training phase, which takes roughly 15 minutes to learn decent tracking performance while the safety filter is assuring it to stay within the safe region. 
The constraint on minimum and maximum altitude derived from the safe set is plotted together. This values vary depending on the system's current state and the level of the value function that the safe set is contracted to. Though there is no distribution shift of the disturbance yet, the contraction happens sometimes due to noise of the disturbance measurement.

Next, we introduce the effect of wind, $d(x)$ whose value is sampled from normal distribution with maximal mean value $3.0\text{m/s}^2$ at $x_1=0.35$m. The drone now faces an \textit{unexpected} disturbance when it approaches the bottom of the reference trajectory. Thus, it contracts its safe set and waits in the conservative safe region until the new safety information is updated. Figure~\ref{fig:verticalflight} shows the process of new safety information updated with our method (top) and recomputed from scratch by using the original method (bottom). The update takes 2.43s for the original method, whereas our method can do this task within 0.71s. Note that in the figure, the difference is minute because most of the time until the safe set is updated is spent on data collection and training the GP model. Once the safety information is updated, it learns that the wind is actually blowing in the direction of ``helping" the drone to stay away from the ground, therefore it is allowed to approach close to the bottom of the trajectory again. Moreover, when disturbance shift is not present, while original method takes an average of 2.60s to recompute the safe set every time it gets new disturbance model, our new method is able to update it in an average of 0.13s, which reduces redundant computation significantly.\vspace{-0.5em}

\subsection{Extension to 10D quadcopter Model}
\label{subsec:10DDemo}

\begin{figure}
\centering
\includegraphics[width=.75\columnwidth]{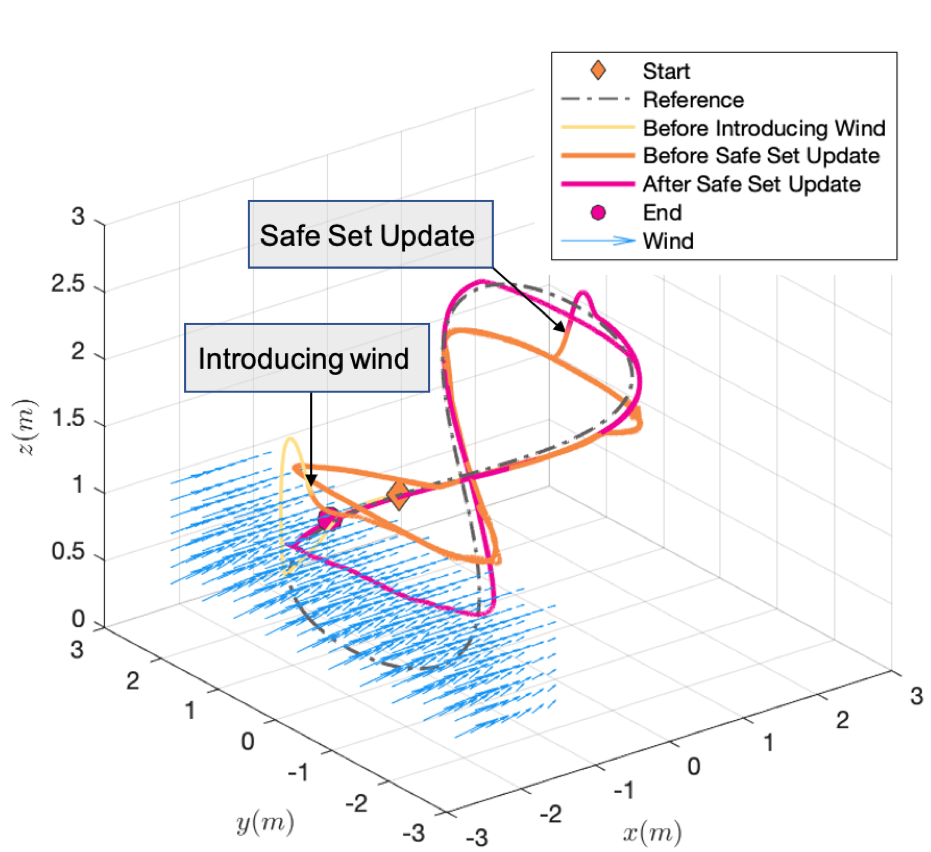}
\caption{Demo of 10D near-hover quadcopter. The reference trajectory is the dashed line. The orange diamond and pink circle denote the start and end of the quadcopter trajectory. The quadcopter begins in yellow, and then experiences a sudden change in wind (blue arrows). Its safe set contracts to a negative level set of the value function (orange trajectory) until the safety analysis update is complete (pink trajectory).}
\label{fig:demo_10D}
\vspace{-2em}
\end{figure}
To demonstrate the scalability of the new framework we test our method with the 10D quadcopter model described in Sec~\ref{sec:new_framework}. The quadcopter must follow a figure-eight reference trajectory in 3D space using an LQR-based performance controller while maintaining safety. The constraint set consists of bounds in $p_x,p_y \in [-2.5, 2.5]$m, $p_z \in [0.35,2.8]$m, velocity $v_x,v_y,v_z \in [-3.5, 3.5]$m/s, and angles $\theta_x,\theta_y \in [-\frac{\pi}{8},\frac{\pi}{8}]$.

The simulation begins without any external disturbances. Shortly thereafter, the wind $[d_x\; d_y \; d_z]$ is introduced in the left bottom corner of the room (see blue arrows in Fig.~\ref{fig:demo_10D}, its quantities are sampled from normal distribution, whose maximal mean norm is 3.9$\text{m/s}^2$ at the edge of the room). The safe set contracts to a negative level set of the initial value function, constraining the quadcopter (orange trajectory). Meanwhile, the new framework computes an updated value function.  Upon completion, the new safe set provides an updated safety guarantee based on the wind disturbance, and the quadcopter continues to track the figure-eight as well as possible while maintaining guaranteed safety (pink). Updating the safety analysis during the simulation took an average of 206.6s (3.3 min). This means a system of greater than three dimensions can now learn and update its safety guarantees online.

\section{Discussion \& Conclusion}
\label{sec:conclusion}

In this paper we used decomposition, warm-starting, and a simple adaptive grid to speed up the computation of safe sets and controllers for autonomous systems. This is particularly useful when the system needs to update its safety analysis online when faced with new information about uncertainties in the environment or system dynamics.  Using our methods we were able to compute and update the safety analysis for a 10D near-hover quadcopter in an average of 3.3 minutes, as opposed to 1 to 2 hours for the prior work to update safety for a 4D system or an intractable amount of time for the 10D system.  This new framework allows learning for safety to be applied to a much larger class of realistic systems.

This work can be extended in several ways. First, the techniques introduced in the new framework can be generalized beyond wind disturbances, updating the safety analysis when changes in knowledge of obstacles, model uncertainty, or other agents occur. The efficiency of this framework can also be sped up through (a) efficient toolbox implementations of HJ reachability, \cite{chen2020optimized_dp, tanabeBEACLS} (b) more sophisticated adaptive gridding, and (c) localized warm-starting \cite{bajcsy2019efficient}. Finally, the code was written to easily generalize to any robot using ROS, and we are eager to perform hardware experiments across different platforms.








\balance
\printbibliography
\end{document}